\PassOptionsToPackage{section}{placeins}

\documentclass[10pt]{cai26}
\usepackage{graphicx}
\usepackage{booktabs}
\usepackage{amsmath,amssymb}
\usepackage{mathtools}
\usepackage{microtype}
\usepackage{enumitem}
\usepackage{multirow}
\usepackage{float}
\usepackage{algorithm}
\usepackage{algpseudocode}
\usepackage{placeins}
\usepackage{listings}

\usepackage[most]{tcolorbox}
\newtcolorbox{promptbox}{
  colback=white,
  colframe=black,
  boxrule=0.4pt,
  arc=2pt,
  left=6pt,right=6pt,top=6pt,bottom=6pt,
  fontupper=\ttfamily\small,
  breakable
}

\newcommand{\ind}{\mathbb{I}}
\newcommand{\Text}{\ensuremath{\mathrm{Text}}}
\newcommand{\Nums}{\ensuremath{\mathrm{Nums}}}

\usepackage{dblfloatfix}          
\usepackage[section]{placeins}    

\usepackage{tabularx}
\usepackage{adjustbox}
\usepackage{makecell}
\usepackage{xurl}
\usepackage{ragged2e}
\newcolumntype{Y}{>{\RaggedRight\arraybackslash}X}

\usepackage{verbatim}

\allowdisplaybreaks
\begin{document}
\def\conferenceyear{2026}
\begin{center}

\title{Decomposing Retrieval Failures in RAG for Long-Document Financial Question Answering}
\maketitle

\thispagestyle{empty}
\pagenumbering{gobble}

\begin{tabular}{cc}
Amine Kobeissi\upstairs{\affilone, \affiltwo, *}, Philippe Langlais \upstairs{\affilone, \affiltwo}
\\[0.25ex]
{\small \upstairs{\affilone} Department of Computer Science and Operations Research, Université de Montréal} \\
{\small \upstairs{\affiltwo} RALI} \\
\end{tabular}
  
\emails{
  \upstairs{*} amine.kobeissi@umontreal.ca 
}\vspace*{0.1in}
\end{center}

\begin{abstract}
Retrieval-augmented generation is increasingly used for financial question answering over long regulatory filings, yet reliability depends on retrieving the exact context needed to justify answers in high stakes settings. We study a frequent failure mode in which the correct document is retrieved but the page or chunk that contains the answer is missed, leading the generator to extrapolate from incomplete context. Despite its practical significance, this within-document retrieval failure mode has received limited systematic attention in the Financial Question Answering (QA) literature. We evaluate retrieval at multiple levels of granularity, document, page, and chunk level, and introduce an oracle based analysis to provide empirical upper bounds on retrieval and generative performance. On a 150 question subset of FinanceBench, we reproduce and compare diverse retrieval strategies including dense, sparse, hybrid, and hierarchical methods with reranking and query reformulation. Across methods, gains in document discovery tend to translate into stronger page recall, yet oracle performance still suggests headroom for page and chunk level retrieval. To target this gap, we introduce a domain fine-tuned page scorer that treats pages as an intermediate retrieval unit between documents and chunks. Unlike prior passage-based hierarchical retrieval, we fine-tune a bi-encoder specifically for page-level relevance on financial filings, exploiting the semantic coherence of pages. Overall, our results demonstrate a significant improvement in page recall and chunk retrieval.
\end{abstract}


\section{Introduction}
Large language models often produce fluent and accurate answers, yet domain specific tasks such as financial question answering require verifiable grounding in long source documents. Retrieval augmented generation (RAG) addresses this by selecting candidate passages from a corpus and generating an answer conditioned on those passages. In this setting, retrieval quality is the main determinant of whether an answer is supported by the correct context.

In practice, retrieval success is not a single concept. A pipeline can retrieve the correct filing and still miss the page or chunk that contains the answer, especially in long reports with repeated templates, dense tables, and similar sections over years. Thus, when the correct chunks are not retrieved, the generator can produce plausible but incorrect responses. Furthermore, document-level metrics can obscure cases where the correct document is retrieved but the answer-bearing context is missed, and generation metrics (answer accuracy, ROUGE-L) do not directly distinguish retrieval from generation failures. For example, low answer accuracy could stem from either retrieving wrong chunks or from generation errors given the correct chunks. We address this with page and chunk retrieval metrics and oracle bounds that separate document discovery from within-document context retrieval

\begin{figure}
    \centering
    \includegraphics[width=0.5\linewidth]{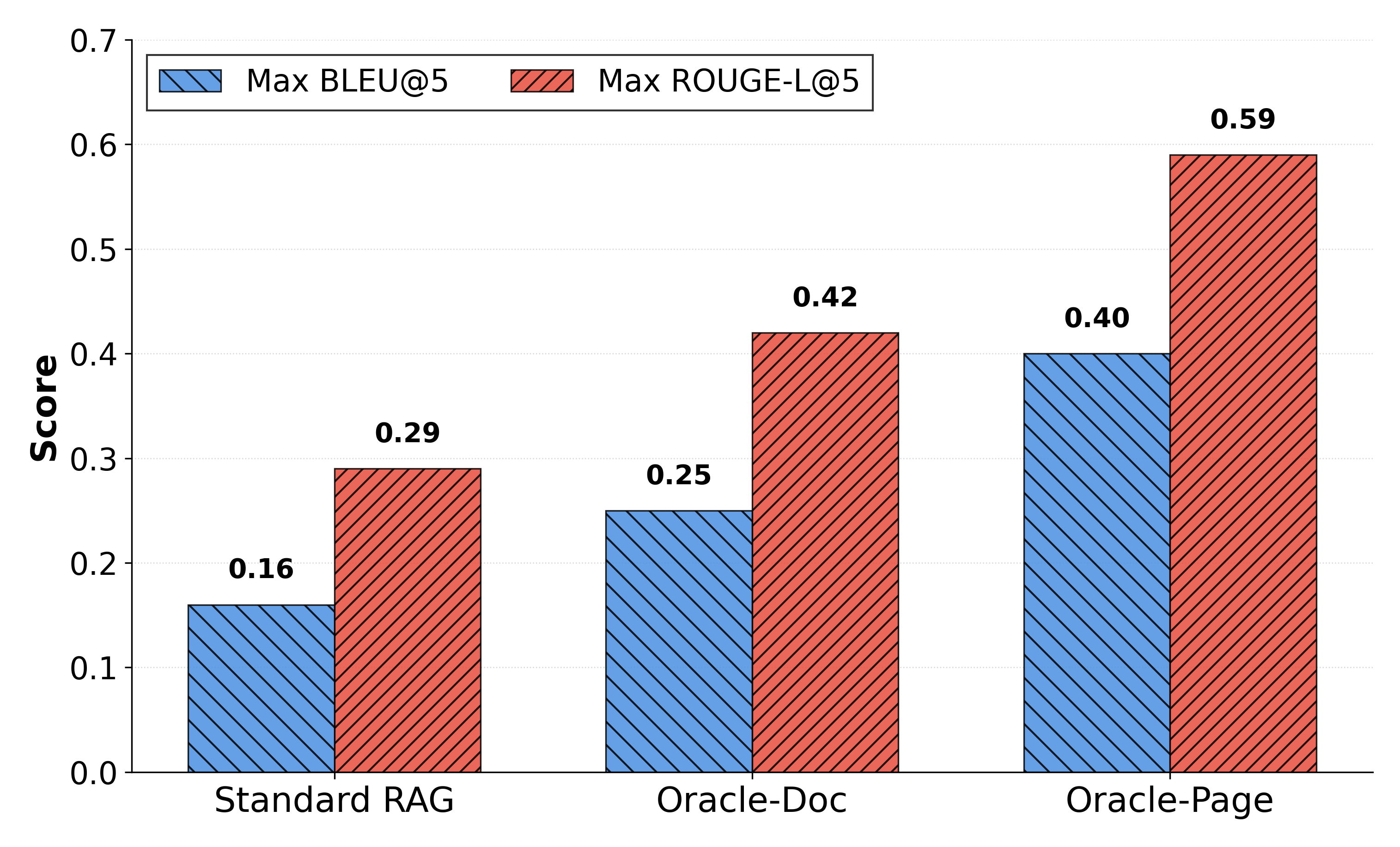}
    \caption{Maximum BLEU and ROUGE-L scores between retrieved chunks and gold chunks under standard retrieval and oracle settings}
    \label{fig:overlap_gap}
\end{figure}

We study this failure mode in financial question answering over U.S. Securities and Exchange Commission (SEC) filings using FinanceBench~\cite{islam2023financebench}. We propose an oracle based analysis that quantifies how much performance is lost due to imperfect document versus imperfect page or chunk retrieval inside the correct filing. The potential benefits of this idea can be visualized in Figure~\ref{fig:overlap_gap}, where we compare retrieval performance in different settings. Since retrieval quality ultimately determines generation quality, we use BLEU ~\cite{papineni2002bleu} and ROUGE-L ~\cite{lin2004rouge} between retrieved chunks and gold chunks as proxies for retrieval performance. We can see that standard RAG pipelines are inadequate compared to oracle settings, revealing the retrieval gap.

\paragraph{Contributions.}
\begin{enumerate}[leftmargin=*,itemsep=2pt]
\item We define an oracle-based retrieval gap analysis that decomposes retrieval into document, page, and within page (chunk) discovery.
\item We provide a comprehensive evaluation of common retrieval strategies on a shared multi document index for FinanceBench, including dense, sparse, hybrid, hierarchical retrieval, query reformulation, and reranking.
\item We introduce a domain fine-tuned page scorer that ranks pages before chunk retrieval, providing a novel approach to close the retrieval gap.
\item We analyze performance by document type, and question type, showing that retrieval difficulty varies across categories.

\end{enumerate}

\section{Related Work}
Retrieval augmented generation combines retrieval with conditional generation to improve factuality and traceability by grounding answers in retrieved text \cite{lewis2020rag}. In long document settings, retrieval quality becomes the main bottleneck because many passages are topically similar while only a small subset contains answer bearing context \cite{izacard2021leveraging}. This motivates evaluations that measure not only whether the correct source document is found, but also whether retrieval locates the optimal page and chunk region needed for a verifiable answer.

Financial question answering is challenging because filings are long, semi structured, and numerically sensitive. Benchmarks such as as FinQA ~\cite{chen2021finqa}, TAT-QA ~\cite{zhu2021tatqa}, and ConvFinQA ~\cite{chen2022convfinqa} emphasize numerical reasoning, often requiring multi-step computation grounded in tables and narrative context. FinanceBench adds gold evidence annotations in SEC filings and supports evidence-based evaluation at scale \cite{islam2023financebench}. Newer retrieval oriented suites such SEC-QA \cite{lai2025secqa} further highlight practical workflows where users require both the answer and the supporting retrieval context.

Retrieval for filings typically relies on chunking plus sparse, dense, or hybrid retrieval, query reformulation, and reranking. Recent work has emphasized the importance of domain-adapted preprocessing and retrieval strategies for financial documents.
Kim et al. ~\cite{kim2025optimizing} propose a three-phase pipeline that combines query expansion, corpus preprocessing, hybrid retrieval with fine-tuned embedders, and a DPO-trained reranker with document selection. Evaluated across several financial QA datasets, their method achieves substantial improvements in retrieval precision. Wang et al. ~\cite{finsage2024} introduce FinSage, a multi-aspect RAG framework that unifies multi-modal preprocessing, multi-path retrieval and a domain-specialized reranker, achieving strong retrieval results and improvements in accuracy over baselines on FinanceBench.
While these systems demonstrate strong end-to-end performance, they do not explicitly decompose retrieval failures into document discovery versus within-document retrieval. Our work complements this by introducing an oracle-based analysis that separately quantifies headroom from imperfect document retrieval and imperfect page or chunk retrieval within the correct filing. We show that while methods like HyDE and reranking improve document discovery, substantial page and chunk retrieval gaps remain. To address this, we introduce a domain fine-tuned page scorer that explicitly addresses granular page level retrieval, providing a diagnostic framework to measure and close the retrieval gap in financial question answering.


\section{Data and Experimental Setup}
\label{sec:oracle_gap}
\subsection{Dataset}
We use the open source subset of FinanceBench with 150 questions. Each sample includes a question, a reference answer, and ground truth annotations that identify the gold document and page numbers. We treat the gold pages as $P^\star$ during evaluation. Table~\ref{tab:example_instance} shows a simplified instance from FinanceBench.

\begin{table}[!htbp]
\centering
\small
\setlength{\tabcolsep}{4pt}
\renewcommand{\arraystretch}{1.08}
\begin{tabularx}{\columnwidth}{lY}
\toprule
\textbf{Field} & \textbf{Value} \\
\midrule
Company & Amcor \\
Document & \texttt{AMCOR\_2020\_10K} \\
Question type & Metrics Generated \\
Reasoning & Information Extraction \\
Question & What is Amcor's year end FY2020 net AR (in USD millions)? Address the question by adopting the perspective of a financial analyst who can only use the details shown within the balance sheet. \\
Answer & \$1616.00 \\
Evidence page & 49 \\
Evidence text & Amcor plc and Subsidiaries Consolidated Balance Sheet (in millions) As of June 30, 2020 2019 Assets Current assets: Cash and cash equivalents \$ 742.6 \$ 601.6 Trade receivables, net  \textbf{1,615.9} 1,864.3  Inventories...\\
\bottomrule
\end{tabularx}
\caption{Instance from FinanceBench. Evidence text is the gold passage for retrieval\protect\footnotemark}
\label{tab:example_instance}
\end{table}
\footnotetext{Evidence text is a passage from the evidence (gold) page containing the answer bearing information.}

FinanceBench includes three question types: domain relevant, metrics generated, and novel generated. Domain relevant questions often require locating narrative statements and may reference multiple periods, and are generic questions relevant to financial analysis of a publicly-traded companies. Metrics generated questions (as in Table \ref{tab:example_instance}) require tight retrieval to the right statement or table to extract base figures for downstream calculation, they require computation and reasoning over different financial figures. Novel generated questions are designed to involve reasoning and are very specific to the company, industry, and report in question. They were designed to require reasoning and not to be purely extractive.

FinanceBench also draws from multiple document types. We report results by document type because retrieval difficulty can differ between annual reports and quarterly reports due to length and section structure. The documents along with their proportion in the data are 10K (74.7\%), 10Q (10\%), Earnings (9.3\%), and 8K (6\%) reports.

\subsection{Task Setup}
\label{sec:task_setup}
\begin{figure}[!htbp]
    \centering
    \includegraphics[width=0.75\linewidth]{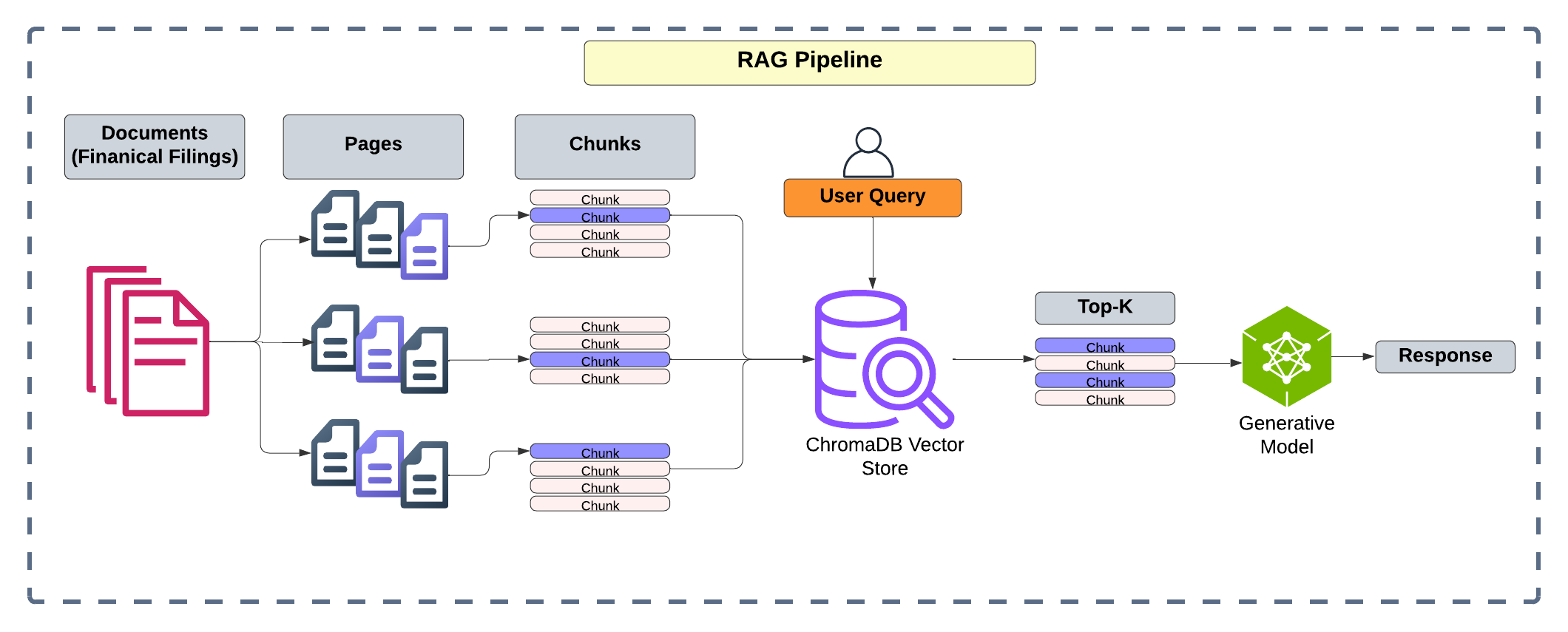}
    \caption{Overview of RAG pipeline. Documents are decomposed into pages and chunks. For a query, blue pages and chunks represent the gold context containing the answer.}
\label{fig:rap_pipeline}
\end{figure}

Figure ~\ref{fig:rap_pipeline} illustrates the overall RAG pipeline evaluated in this work. Let $\mathcal{D} = \{d_1, \ldots, d_D\}$ be a set of documents. Each document $d\in \mathcal{D}$ contains $n$ pages $\{p_1^d, \ldots, p_{n}^d\}$, we extract the page from the PDF document using the FinanceBench preprocessing\footnote{https://github.com/patronus-ai/financebench}. We build a chunk corpus $\mathcal{C}$ by splitting page text into overlapping spans, where the text is of 1024 tokens and overlap of 128 tokens\footnote{Following Yepes et al. \cite{jimenoyepes2024financial_chunking}, and our ablations, we find that larger chunk sizes aid in Financial RAG}. Each chunk is a tuple $(c, d, p)$ with chunk text $c$, document id $d$, and a 0-indexed page number $p$. For a query $q$, a retriever produces a ranked list of chunks,
\begin{equation*}
\mathcal{R}_k(q) = \operatorname{Top}{k}_{(c,d,p) \in \mathcal{C}} \; s(q, c) = \{(c_i, d_i, p_i)\}_{i=1}^{k},
\end{equation*} 

where $s$ is a scoring function. The generator produces an answer $\hat{y} = G_\theta(q, \mathcal{R}_k(q))$. FinanceBench provides a gold document $d^\star$ and gold pages $P^\star$ for each query.

\subsection{Oracle Retrieval Conditions}
We define three retrieval settings that differ by candidate restriction. First, standard retrieval, where we retrieve over the entire search space. Second, Oracle document retrieval, where the candidates in the search space are restricted to the gold filing document, $\mathcal{C}^{\mathrm{doc}}(d^\star) = \{(c,d,p) \in \mathcal{C} \mid d = d^\star\}$. Third, Oracle page retrieval, where candidates are restricted to the gold filing and gold pages, $\mathcal{C}^{\mathrm{page}}(d^\star, P^\star) = \{(c,d,p) \in \mathcal{C} \mid d = d^\star,\; p \in P^\star\}$, resulting in a near complete reduction of noise.

These conditions provide empirical upper bounds. The Oracle document setting quantifies headroom due to imperfect page and chunk discovery. The Oracle page setting quantifies headroom due to imperfect chunk retrieval when the gold pages and documents are known. Evaluating both bounds helps isolate where errors arise. More broadly, oracle bounds provide a common reference across systems. The gap to oracle normalizes for pipeline differences and contextualizes raw scores.

\subsection{Evaluation Metrics}
\label{sec_metrics}

For each question $q$, FinanceBench provides a gold filing $d^\star(q)$ and a set of gold pages $P^\star(q)$. Let the top $k$ retrieved chunks be
$\mathcal{R}_k(q) = \{(c_i, d_i, p_i)\}_{i=1}^{k}$,
where $d_i$ and $p_i$ denote the filing id and page index of chunk $c_i$.

First, we measure document recall. Since each question has a single gold filing, document recall at $k$ is equivalent to hit,
\begin{equation*}
\DocRec@k(q) = \ind\!\left[d^\star(q) \in \{d_i\}_{i=1}^{k}\right].
\end{equation*}

Then, we measure page recall at $k$, where there can be many gold pages.
\begin{equation*}
\PageRec@k(q) =
\frac{\left|P^\star(q) \cap \{p_i \mid d_i = d^\star(q)\}_{i=1}^{k}\right|}{\left|P^\star(q)\right|}.
\end{equation*}

Finally, as a chunk-level proxy, we compute the maximum ROUGE-L~\cite{lin2004rouge} and BLEU~\cite{papineni2002bleu} similarity between any retrieved chunk and the gold evidence concatenated into a single reference text, $e^\star(q)$. Given the top-$k$ retrieved chunks $\{c_i\}_{i=1}^{k}$, for each query we report
\begin{equation*}
\mathrm{CtxROUGE}\text{-L}@k(q) =
\max_{i \in \{1,\dots,k\}} \mathrm{ROUGE}\text{-L}\!\left(c_i, e^\star(q)\right),
\end{equation*}
and analogously $\mathrm{CtxBLEU}@k(q)$ using BLEU. To evaluate the generator, ROUGE-L scores on the generated output against the reference answer are used along with a numeric match for metric based questions. To calculate the numeric match, given a reference answer $y^\star$ and a model prediction $\hat{y}$, we extract sets of numbers
$R=\Nums(y^\star)$ and $P=\Nums(\hat{y})$ using a regular expression after removing commas and currency symbols.
We count an example as a numeric match if there exist $r \in R$ and $p \in P$ such that $r$ and $p$ are close
under \texttt{numpy.isclose}\footnote{https://numpy.org/doc/stable/reference/generated/numpy.isclose.html},
using $\texttt{atol}=0.03$ and $\texttt{rtol}=0.03$.

We report macro averages over all 150 questions and use $k=5$ throughout the evaluation, other than for numeric match which is only reported on the 50 metrics generated questions.

\subsection{Generation Setup}
The generator model, prompt template, and decoding settings are fixed across all retrieval methods. We use Qwen-2.5-7B-Instruct as our generator model. The prompt template used can be found in Appendix \ref{app:gen_tem}.

\FloatBarrier

\section{Methodology and Approach}
\label{sec:methodology}

We evaluate a suite of retrieval strategies designed to improve document, page, and chunk level discovery. All strategies share the same task setup as described in section \ref{sec:task_setup}. For dense methods we use a shared \texttt{ChromaDB}\footnote{https://github.com/chroma-core/chroma} vector store built once per configuration. The implementation is model agnostic and supports embedding backbones such as \texttt{BGE-M3} ~\cite{chen2024bgem3}. Chunks are stored in a single \texttt{ChromaDB} collection, and retrieval uses \texttt{similarity\_search}.

\subsection{Dense Retrieval}
Dense retrieval maps a query and each chunk to a shared vector space \cite{karpukhin2020dpr} and ranks chunks by vector similarity. Let $E(\cdot)$ be the embedding model and $c$ a chunk. We compute a query embedding $E(q)$ and retrieve the top $k$ chunks by cosine similarity. All dense variants differ only in the choice of embedding backbone while sharing the same vector store, chunk size and overlap, and chunk metadata.

\subsection{Sparse retrieval}
Sparse retrieval excels at keyword matching and interpretability, thus offering potential advantages for Financial QA. We test BM25~\cite{robertson2009probabilistic} as a sparse baseline and SPLADE~\cite{formal2021splade} as a learned sparse retriever that produces sparse term weighted vectors. BM25 is implemented using \texttt{rank-bm25}\footnote{https://github.com/dorianbrown/rank\_bm25}. Tokenization uses a finance oriented preprocessing function that preserves currency symbols, percents, 
and financial tokens such as numbers with commas and decimals. The retriever returns the top $k$ chunks with the highest BM25 scores. SPLADE uses the checkpoint \texttt{naver/splade-cocondenser-ensembledistil}. Due to architectural constraints of the checkpoint, we truncate inputs to 512 tokens, and we keep the top $N=256$ weighted terms for efficiency. The token truncation may impact SPLADE results since it sees less context per chunk than methods with 1024 tokens. Retrieval ranks chunks by sparse dot product between query and chunk vectors.

\subsection{Hybrid Fusion}
We combine dense and sparse ranked lists using Reciprocal Rank Fusion \cite{cormack2009rrf}. Let $\text{rank}_r(c)$ be the rank of chunk $c$ under retriever $r$. Then RRF assigns
\begin{equation*}
s_{\text{RRF}}(c) =
\sum_{r \in \mathcal{M}} \frac{1}{k_{\text{rrf}} + \text{rank}_r(c)},
\end{equation*}
where $k_{\text{rrf}}=60$ and method are equally weighted. To fuse and compare outputs robustly across systems, we align chunks using stable SHA1 hashes computed from normalized chunk text concatenated with metadata fields for source and page. This creates consistent keys even when different retrievers produce results through different internal data structures.

\subsection{Query Reformulation, HyDE and Multi-HyDE}
Lexical mismatch between short questions and filing language is common, particularly when the question uses conversational phrasing while filings use formal accounting terminology. To address this challenge, we attempt three approaches. First, we apply a deterministic query expansion baseline for retrieval, while generation uses the original question. We expand common finance acronyms by appending an expanded version, for example CAPEX becomes capital expenditure, and PPE becomes property plant equipment. We also normalize temporal expressions such as FY2018, or FY18 into a canonical form fiscal year 2018.

For the other two methods, we follow HyDE and its multi sample variant proposed in the literature \cite{gao2023hyde,srinivasan2025multihyde}. HyDE generates a hypothetical passage intended to resemble the target context, then retrieves using the embedding of that passage rather than the raw query embedding. We use Qwen-2.5-7B-Instruct as our LLM to generate the hypothetical passages. We set temperature to 0.7 to encourage diversity in the outputs and cap the generated passage to 200 tokens.
 The prompt uses the following template,
\begin{promptbox}
Please write a short financial report excerpt (1 paragraph) that precisely answers the question below. Do not introduce the text, just write the excerpt.

Question: \{question\}

Passage:
\end{promptbox}

For Multi-HyDE we use the same set up as HyDE but generate 4 hypothetical passages $\{h_i\}_{i=1}^4$, embed each passage, then form a single query vector $\bar{h}$ to perform the search,  
\begin{equation*}
\bar{h} = \frac{1}{4}\sum_{i=1}^4 E(h_i).
\end{equation*}

\subsection{Hierarchical Retrieval}
Hierarchical retrieval addresses long context by retrieving at a coarse granularity and then refining at a finer granularity within the retrieved region.  Recent work targets fine grained retrieval in financial filings via a hierarchical approach, often taking the form of document retrieval followed by passage retrieval, with additional refinement aggregation~\cite{izacard2021leveraging,ma2025hirec,li2025fingear}. We implement a simple Parent-Child style hierarchy that retrieves child chunks at a fine granularity, then maps to larger parent chunks for generation context. We index small child chunks for retrieval and maintain a deterministic mapping from each child to a larger parent span. After retrieving children, we deduplicate by parent id and return up to $k$ unique parents. A parent score is derived from its children by taking the best similarity score among children assigned to that parent, then ranking parents accordingly.

\subsection{Cross-Encoder Reranking}
In long filings, reranking can help separate near duplicate regions and improve retrieved context ordering after a high recall first stage, which is important when multiple sections contain similar language across years or business segments~\cite{li2025fingear}. We add a second stage reranker using \texttt{BAAI/bge-reranker-v2-m3}. For each query we first retrieve $N=20$ candidates from a base retriever, then rerank with the cross-encoder and keep the top $k$.

\subsection{Page-then-Chunk Retrieval}
We propose a hierarchical retrieval method that identifies relevant pages and then retrieves chunks only from those pages, reducing noise from irrelevant sections while preserving fine-grained context retrieval. To do this, we fine-tune a bi-encoder to identify optimal pages while enabling efficient inference through pre-computed page embeddings. The method has two stages. Stage one ranks pages in the corpus and returns the top $P$ pages. Stage two runs a retriever over chunks whose metadata matches the selected pages.

For each filing $d$, we extract the text of each page $p$ and apply a deterministic normalization function $N(\cdot)$ that collapses whitespace and truncates long pages to a fixed length. We set the maximum length to $M = 2000$ characters. The learned page scorer replaces a pretrained page encoder with a fine-tuned bi-encoder. Let $\Text(d,p)$ be the page text of a document, we compute their embeddings from $N(\Text(d,p))$. For the scoring function, we train a bi-encoder $E_{\theta_p}$ initialized from \texttt{BAAI/BGE-M3} that maps queries and pages into a shared vector space. The learned page relevance score is the cosine similarity of the learned embedding of the page and query, $s_{\text{page}}(q,d,p)
= \cos\!\Big(E_{\theta_p}(q),\, E_{\theta_p}\big(N\Text(d,p)\big)\Big).
\label{eq:page_score_learned}$ At inference, pages are ranked by $s_{\text{page}}$ and the top $P$ pages are passed to the chunk stage.

For each question $q$, FinanceBench provides a gold filing $d^\star(q)$ and a set of gold page indices $P^\star(q)$. To train, we construct positives by pairing each question with every annotated page in its gold filing,
$\mathcal{S}
= \{(q,\,(d^\star(q),p^+)) \mid p^+ \in P^\star(q)\}.
\label{eq:page_pairs}
$ During training, we sample minibatches from the training subset of 
$\mathcal{S}$ and apply contrastive learning\footnote{https://huggingface.co/blog/dragonkue/mitigating-false-negatives-in-retriever-training} to fine-tune the embedding model $E_{\theta_p}$ 
using the Multiple Negatives Ranking loss with in-batch negatives ~\cite{henderson2017smartreply}. For a minibatch of size 
$B$ with positive pairs $\{(q_i,(d_i^\star,p_i^+))\}_{i=1}^B$ sampled from $\mathcal{S}$, the loss is

\begin{equation*}
\mathcal{L}_{\text{MNR}}
= -\frac{1}{B}\sum_{i=1}^{B}
\left[
s_{\text{page}}(q_i,d_i^\star,p_i^+)
- \log \sum_{j=1}^{B} \exp\big(s_{\text{page}}(q_i,d_j^\star,p_j^+)\big)
\right].
\label{eq:mnr}
\end{equation*}
For each $q_i$, its paired gold page is the positive target, and all other pages in the minibatch are negatives.

We train\footnote{Experiments were run on a RTX 3090 GPU with 32GB memory.} the page scorer using document-level splits to prevent data leakage, ensuring no document appears in both training and evaluation sets. We perform 5-fold cross-validation, partitioning the questions by their source documents. For each fold, we train on 80\%  and evaluate on the remaining 20\% of documents. Training uses a learning rate of 2e-5, batch size of 16, and 15 epochs. Final metrics are aggregated across 
all folds.

\subsubsection{Inference and Integration with Chunk Retrieval}
After training, we embed every page in the corpus with $E_{\theta_p}$ and index them with $(d,p)$ metadata. At inference, given query $q$, we retrieve the top $P=20$ pages (via ablation in Appendix~\ref{app:P}) under $s_{\text{page}}$, filter chunks to only those from selected pages, then apply chunk-level retrieval to select the final top $k$ chunks for generation. This targets page-level relevance while preserving fine-grained chunk selection. Algorithm \ref{alg:page_then_chunk} provides the routine.
\begin{algorithm}[!htbp]
\caption{Page-then-chunk retrieval using fine tuned page scorer}
\label{alg:page_then_chunk}
\begin{algorithmic}[1]
\Require Query $q$, Fine tuned page encoder $E_{\theta_p}$, Chunk encoder $E_{\text{chunk}}$, $P$ pages, $k$ chunks
\Require Page index $\mathcal{I}_{\text{page}}$ storing entries $\big((d,p),\, v^{\text{page}}_{d,p}\big)$
\Statex \hspace{\algorithmicindent} where $v^{\text{page}}_{d,p} = E_{\theta_p}\!\big(N(\Text(d,p))\big)$ is computed offline once
\Require Chunk index $\mathcal{I}_{\text{chunk}}$ storing entries $\big(c,\, v^{\text{chunk}}_{c},\, d_c,\, p_c\big)$
\Statex \hspace{\algorithmicindent} where $v^{\text{chunk}}_{c} = E_{\text{chunk}}(c)$ is computed offline once

\State \textbf{Stage 1: Page retrieval}
\State $e_q^{\text{page}} \gets E_{\theta_p}(q)$
\State $\mathcal{P}_P(q) \gets \textsc{Search}_{\cos}\!\left(\mathcal{I}_{\text{page}},\, e_q^{\text{page}},\, P\right)$
\Comment{Nearest neighbor search over stored $v^{\text{page}}_{d,p}$}

\State \textbf{Stage 2: Chunk retrieval on filtered search space}
\State $e_q^{\text{chunk}} \gets E_{\text{chunk}}(q)$
\State $\mathcal{C}_{\text{filt}} \gets \{c \in \mathcal{I}_{\text{chunk}} \mid (d_c,p_c) \in \mathcal{P}_P(q)\}$
\State $\mathcal{R}_k(q) \gets \textsc{Search}_{\cos}\!\left(\mathcal{C}_{\text{filt}},\, e_q^{\text{chunk}},\, k\right)$
\Comment{Search over stored $v^{\text{chunk}}_{c}$ in the filtered pool}
\State \Return $\mathcal{R}_k(q)$
\end{algorithmic}
\end{algorithm}

\FloatBarrier

\section{Results and Analysis}
\label{sec:results}

\subsection{Oracle Results}
Table~\ref{tab:oracle_bounds} reports oracle document and oracle page results at $k=5$ using \texttt{BAAI/BGE-M3}. Under the oracle document setting, we have a page recall of 0.60, as well as a BLEU score of 0.25 and ROUGE-L of 0.42. When restricting the search space in the oracle page experiment, the BLEU and ROUGE-L scores increase, meaning that we are retrieving better context. The numeric answer accuracy also increases under the oracle page setting, indicating expectedly that down stream generation improves when the correct page is located. These results provide empirical upper bounds for the remaining experiments.

\begin{table}[!htbp]
\centering
\small
\setlength{\tabcolsep}{4pt}
\renewcommand{\arraystretch}{1.12}
\begin{adjustbox}{max width=\columnwidth}
\begin{tabular}{lrrrrr}
\toprule
Setting &
\DocRec@5 &
\PageRec@5 &
Max BLEU@5 &
Max ROUGE-L@5 &
Numeric Match \\
\midrule
Oracle document & 1.00 & 0.60 & 0.25 & 0.42 & 0.26 \\
Oracle page     & 1.00 &  1.00 & 0.40 & 0.59 & 0.34 \\
\bottomrule
\end{tabular}
\end{adjustbox}
\caption{Oracle results. Evaluation metrics defined in section \ref{sec_metrics}.}
\label{tab:oracle_bounds}
\end{table}

\subsection{Retrieval Results}
We report aggregated retrieval performance on all 150 questions in Table~\ref{tab:main_results}. 
Across all methods, page recall substantially lags document recall, revealing that while baseline methods successfully identify the correct filing, 
they struggle to find answer-bearing context within it. Consistent with prior work, dense 
retrieval outperforms sparse methods on FinanceBench ~\cite{kim2025optimizing}, likely due to lexical mismatch between queries and formal filing language. Among baselines, BGE-M3 combined with Multi-HyDE and reranking performs best (0.46 page recall), showing that query reformulation and cross-encoder scoring effectively improve within-document retrieval. However, in terms of page recall, this method is still 0.14 away 
from the Oracle Document upper bound of 0.60, demonstrating the retrieval gap. Our learned page scorer achieves 0.55 page recall, outperforming baselines and closing the page recall gap to 0.05 for the oracle document setting. We also achieve the highest chunk-level BLEU (0.33) and ROUGE-L (0.46) scores, outperforming all baselines and Oracle Document. This validates that page-level filtering improves retrieval quality over direct chunk search.
\begin{table*}[!htbp]
\centering
\small
\setlength{\tabcolsep}{3pt}
\renewcommand{\arraystretch}{1.12}
\begin{adjustbox}{max width=\textwidth}
\begin{tabular}{l r r r r}
\toprule
Method & 
\DocRec@5 & \PageRec@5 &
Max BLEU@5 & Max ROUGE-L@5 \\
\midrule
Dense baseline BGE-M3   & 0.88  & 0.34  & 0.26 & 0.35  \\
BM25                 & 0.32 & 0.07  & 0.04 & 0.12  \\
SPLADE               & 0.50 & 0.16 & 0.18 & 0.29  \\
Hybrid fusion (BM25 + BGE-M3)        & 0.61 & 0.23 & 0.13 & 0.27  \\
Parent-Child         & 0.31 & 0.13 & 0.12 & 0.22  \\
Query expansion      & 0.68 & 0.27 & 0.08 & 0.24 \\
HyDE                 & 0.86 & 0.40 & 0.25 & 0.37 \\
Multi-HyDE           & 0.85 & 0.42 & 0.27 & 0.39  \\
BGE-M3 + ReRanker & 0.87 & 0.41 & 0.19 & 0.34 \\

BGE-M3 + Multi-HyDE + ReRanker  & 0.93 & 0.46 & 0.28 & 0.40 \\
\midrule
Learned page scorer (Ours) &  \textbf{0.95} & \textbf{0.55} & \textbf{0.33} & \textbf{0.46}\\
\bottomrule
\end{tabular}
\end{adjustbox}
\caption{Aggregated retrieval performance over 150 FinanceBench samples at $k=5$.}
\label{tab:main_results}
\end{table*}

\subsection{Breakdown by Question and Document Type}
We report page and document recall by question type. 
Figure 3 demonstrates that performance varies significantly by question type. On metrics questions, our learned page scorer with a page recall of 0.81 exceeds Oracle Document page recall of 0.68, despite both achieving perfect document recall. This occurs because metrics questions target financial tables and numerical statements that appear in structurally predictable sections, allowing the domain-tuned page embeddings to effectively retrieve the correct page. Conversely, domain and novel questions show larger gaps to their oracle bounds, as these require narrative context that is harder to capture and face different retrieval challenges.

\begin{figure}[!htbp]
        \centering
        \includegraphics[width=0.85\linewidth]{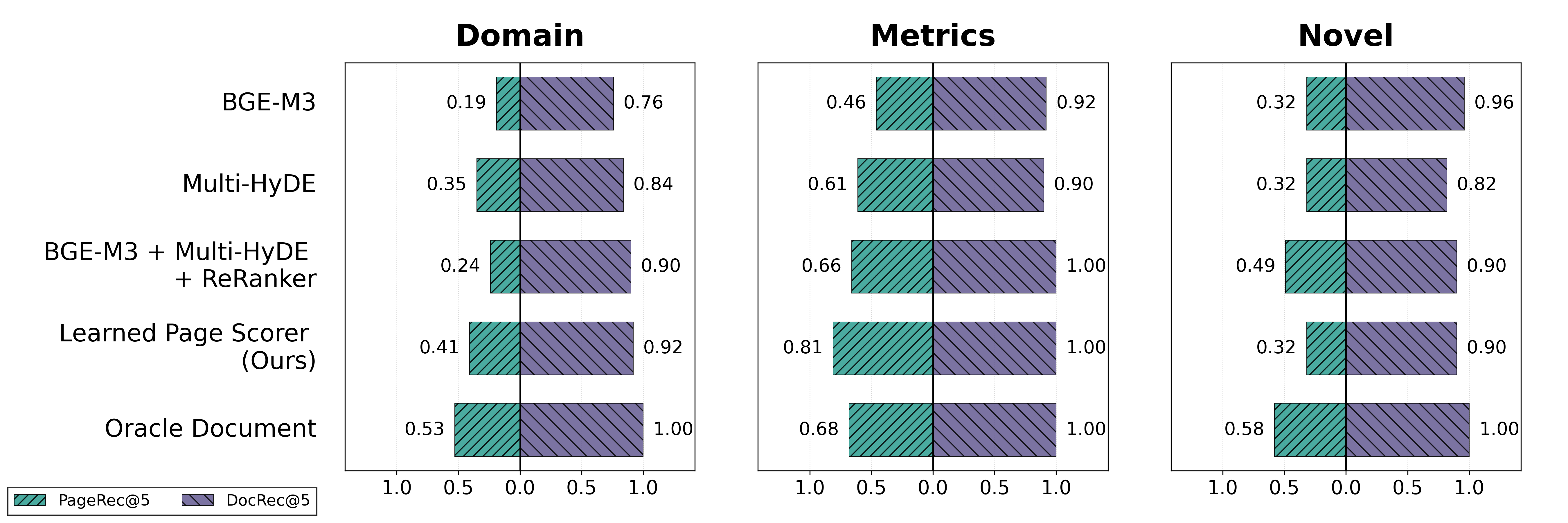}
        \caption{Document and page recall at $k=5$ by question type. 50 questions per type}
        \label{fig:placeholder}
    \end{figure}

As shown in Table ~\ref{tab:by_doc}, performance varies substantially by filing type. On 10K filings, which comprise 
75\% of our dataset, the learned page scorer achieves 0.62 page recall, exceeding both the 
best baseline (0.45) and Oracle Document (0.56). The strong performance on 10Ks is likely due to the sample size in the training data and reflects the consistent structure and length of annual reports. In contrast, performance on Earnings 
call (EC)  transcripts is weak (0.10), falling far short of the oracle bound (0.64) and baseline (0.36). Earnings calls have less standardized structure and evidence may appear in conversational Q\&A sections that differ from the training data. Performance on 10Q and 8K filings also underperformed the baseline, likely due to limited representation of these document types in training data.

\begin{table}[!htbp]
\centering
\small
\setlength{\tabcolsep}{3pt}
\renewcommand{\arraystretch}{1.12}
\begin{adjustbox}{max width=\columnwidth}
\begin{tabular}{l rr rr rr rr}
\toprule
& \multicolumn{2}{c}{10K (n=112)} & \multicolumn{2}{c}{10Q (n=15)} & \multicolumn{2}{c}{8K (n=9)} &  \multicolumn{2}{c}{EC (n=14)} \\
\cmidrule(lr){2-3} \cmidrule(lr){4-5} \cmidrule(lr){6-7}  \cmidrule(lr){8-9}
Method & D@5 & P@5 & D@5 & P@5 & D@5 & P@5 & D@5 & P@5 \\
\midrule
BGE-M3 + Multi-HyDE + ReRanker &  0.96  & 0.45 & 0.87 & 0.47 & 0.89 & 0.78 & 0.86 & 0.36 \\
Learned Page Scorer (Ours) & 0.97 & 0.62 & 0.87 & 0.40 & 0.89 & 0.56 & 0.86 & 0.10 \\
\midrule
Oracle Document & 1.00 & 0.56 & 1.00 & 0.60 & 1.00 & 0.89 & 1.00 & 0.64 \\
\bottomrule
\end{tabular}
\end{adjustbox}
\caption{Document (D@5) and page (P@5) recall at $k=5$ by filing type. EC is Earnings.}
\label{tab:by_doc}
\end{table}

\FloatBarrier

\subsection{Generative Results}
Despite improvements in page recall, it is important to evaluate whether downstream 
generation benefits from better evidence retrieval. Table ~\ref{tab:gen_res} shows answer quality using 
Qwen-2.5-7B-Instruct. Our learned page scorer achieves 0.30 numeric match accuracy, 
outperforming the best baseline (12\%) and oracle document (26\%), while also improving 
ROUGE-L scores over strongest baseline (0.15 vs 0.07), supporting our results from Figure ~\ref{fig:placeholder}, where our method outperformed the Oracle Document setting on metrics question. This suggests that page-level filtering successfully targets pages containing the numerical evidence needed for accurate calculation, while 
maintaining answer quality across both extractive and reasoning questions. 

\begin{table}[!htbp]
\centering
\small
\setlength{\tabcolsep}{4pt}
\renewcommand{\arraystretch}{1.12}
\begin{adjustbox}{max width=\columnwidth}
\begin{tabular}{lrr}
\toprule
Method & ROUGE-L & Numeric Match \\
\midrule
BGE-M3 + Multi-HyDE + ReRanker  & 0.07 & 0.12 \\
Learned Page Scorer (Ours) & 0.15 & 0.30 \\
\midrule
Oracle Doc & 0.16 & 0.26 \\
Oracle Page  & 0.20 & 0.34 \\
\bottomrule
\end{tabular}
\end{adjustbox}
\caption{Generative results. Numeric match follows tolerance from FinanceBench}
\label{tab:gen_res}
\end{table}



\section{Conclusion}
We introduced an oracle-based framework to decompose retrieval failures in financial question answering, separating document discovery from page and chunk level retrieval within the correct filing. Across diverse retrieval strategies on FinanceBench, we find that while current methods improve document recall, page-level retrieval gaps remain even when the correct document is retrieved. 
To address this bottleneck, we introduce a page scorer that ranks pages within a document and then restricts chunk retrieval to the top ranked pages. Our approach achieves 55\% page recall, outperforming all baseline methods as well as Oracle Document retrieval on metrics generated questions. This shows that explicit page-level modeling with domain-specific training improves within-document retrieval more effectively than direct chunk matching, even with perfect 
document discovery.

Our study evaluates 150 questions from FinanceBench, predominantly 10-K filings, limiting generalization to other document types or domains. This motivates the need to evaluate our approach on other benchmarks or the full FinanceBench dataset. The page scorer requires gold annotations and relies on random in-batch 
negatives during training, in the future hard negative mining could 
improve results. Answer evaluation uses ROUGE-L and numeric matching, which may not fully capture correctness in financial contexts. Finally, generative results are based on one LLM, using larger or different models may impact generative results.

Promising directions include combining page scoring with reranking or query augmentation, reinforcement learning for iterative retrieval, incorporating document structure metadata, and extending to 
multi-document questions requiring cross-filing context aggregation.

\appendix

\section{Generator Prompt Template}
\label{app:gen_tem}
\begin{promptbox}
    System: You are a financial analyst assistant. Your task is to provide accurate, 
concise, and well-supported answers to questions based on provided financial 
document segments.

Context:
\{retrieved\_evidence\}

Question:
\{query\}

Answer:
\end{promptbox}

\section{Learned Page Scorer Performance by P}
\label{app:P}

\begin{table}[!htbp]
\centering
\setlength{\tabcolsep}{4pt}
\renewcommand{\arraystretch}{1.12}
\begin{adjustbox}{max width=\columnwidth}
\begin{tabular}{r r r r r}
\toprule
P & 
\DocRec@5 & \PageRec@5 &
Max BLEU@5 & Max ROUGE-L@5 \\
\midrule
5   & 0.92 & 0.48 & 0.30 & 0.44 \\
10 & 0.95 & 0.48 & 0.31 & 0.44 \\
20 & 0.95 & 0.55 & 0.33 & 0.46 \\
\bottomrule
\end{tabular}
\end{adjustbox}
\caption{P value ablation for learned
page scorer}
\label{tab:chunk_ablation}
\end{table}

\printbibliography[heading=subbibintoc]

\end{document}